\documentclass[10pt,twocolumn,letterpaper]{article}

\usepackage{iccv}
\usepackage{times}
\usepackage{epsfig}
\usepackage{graphicx}
\usepackage{amsmath}
\usepackage{amssymb}
\usepackage[]{booktabs}
\usepackage[ruled,linesnumbered]{algorithm2e}
\usepackage{url}

\newcommand{\sto}{{\to}}
\newcommand{\slangle}{{<}}
\newcommand{\srangle}{{>}}

\newcommand{\sminus}{{-}}

\usepackage[pagebackref=true,breaklinks=true,letterpaper=true,colorlinks,bookmarks=false]{hyperref}

\iccvfinalcopy 

\def\iccvPaperID{1273} 

\ificcvfinal\fi

\begin{document}

\title{Human Re-identification by Matching Compositional Template  with \\ Cluster Sampling}

\author{Yuanlu Xu$^1$ \quad \quad Liang Lin$^1$\thanks{\small Corresponding author is Liang Lin. This work was supported by Program for New Century Excellent Talents in University (no.NCET-12-0569), Program of Guangzhou Zhujiang Star of Science and Technology (no. 2013J2200067), the Guangdong Science and Technology Program (no. 2012B031500006), and the Guangdong Natural Science Foundation (no. S2013050014548).  } \quad  \quad Wei-Shi Zheng$^1$ \quad \quad Xiaobai Liu$^2$\\
$^1$Sun Yat-Sen University, China  \quad \quad  $^2$University of California, Los Angeles\\
{\tt\small merayxu@gmail.com, \{linliang, wszheng\}@ieee.org, \tt\small lxb@ucla.edu}
}

\maketitle



\begin{abstract}
\vspace{-3mm}

    This paper aims at a newly raising task in visual surveillance: re-identifying people at a distance by matching body information, given several reference examples.  Most of existing works solve this task by matching a reference template with the target individual, but often suffer from large human appearance variability (e.g. different poses/views, illumination) and high false positives in matching caused by conjunctions, occlusions or surrounding clutters.  Addressing these problems, we construct a simple yet expressive template from a few reference images of a certain individual, which represents the body as an articulated assembly of compositional and alternative parts, and propose an effective matching algorithm with cluster sampling.  This algorithm is designed within a candidacy graph whose vertices are matching candidates (i.e. a pair of source and target body parts), and iterates in two steps for convergence. (i) It generates possible partial matches based on compatible and competitive relations among body parts. (ii) It confirms the partial matches to generate a new matching solution, which is accepted by the Markov Chain Monte Carlo (MCMC) mechanism. In the experiments, we demonstrate the superior performance of our approach on three public databases compared to existing methods. 

\end{abstract}
\vspace{-1mm}

\vspace{-3mm}
\section{Introduction} \label{sec:Intro}
\vspace{-2mm}

\begin{figure}[ptb]
\vspace{-1mm}
\begin{center}
\includegraphics[width=3.2in]{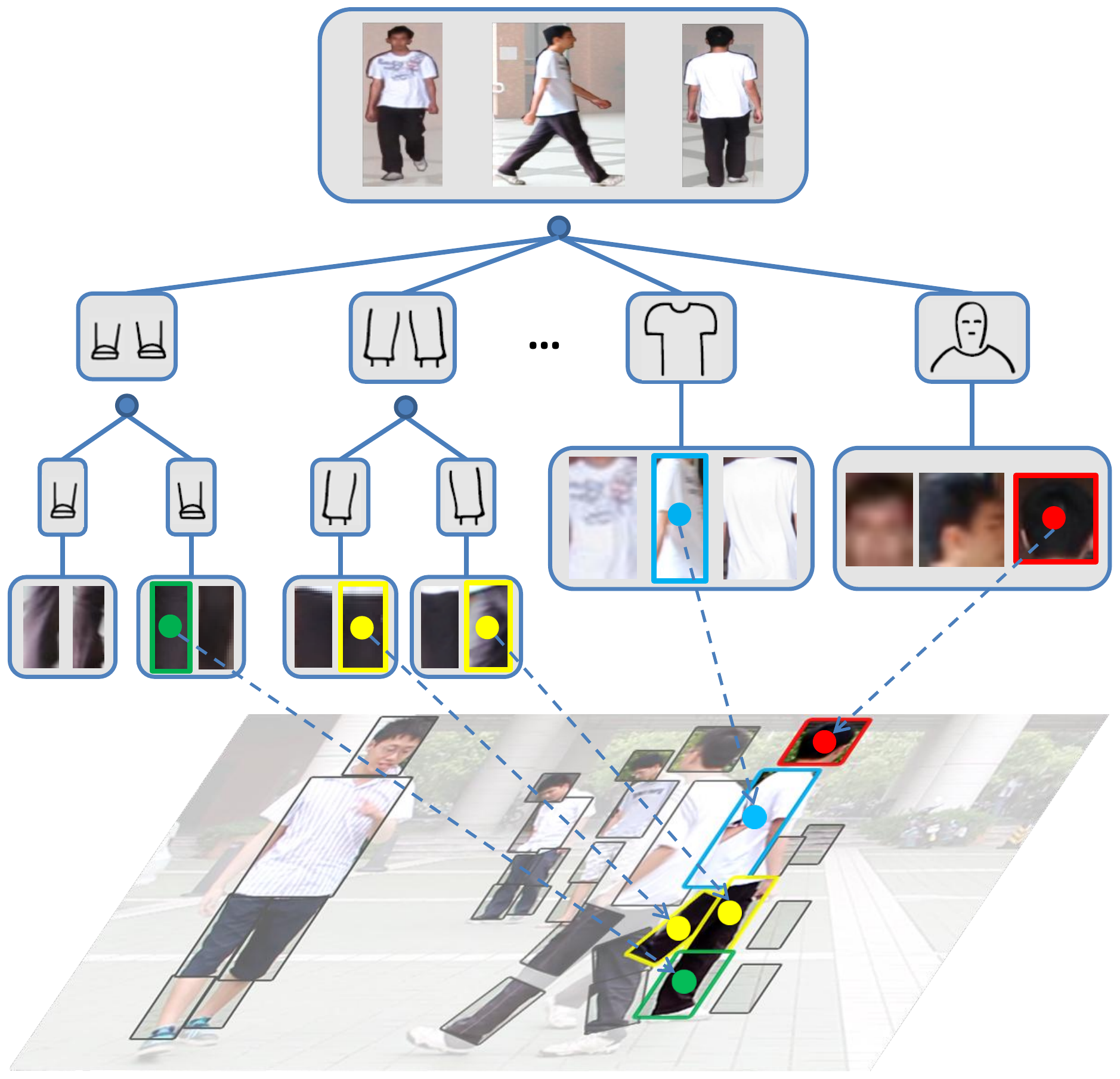}
\end{center}
\vspace{-4mm}
\caption{An illustration of the proposed approach. A query individual is represented as a compositional part-based template, and part proposals are extracted from multiple instances at each parts. Human re-identification is thus posed as compositional template matching. Note certain parts are omitted for clear specification.}
\label{fig:framework}
\vspace{-4mm}
\end{figure}



Person re-identification at a distance increasingly receives attention in video surveillance, particularly for the applications restricting the use of face recognition. But this task is very challenging due to the following difficulties,

{\small \textbullet}\; Robust human representation (signature). There are large variations for human body in appearance, (e.g., different views, poses, lighting conditions). It is usually intractable to construct a template of the individual to be recognized by extracting only low-level image features.

{\small \textbullet}\; Effective human matching (localizing). Given the template, re-identifying targets with the global body information often suffers from high matching false positives, as the targets are possibly occluded or conjuncted with others and backgrounds in realistic surveillance applications. Furthermore, it is desired to accurately localize human body parts in general.

The objective of human re-identification in this work is to recognize an individual by employing body information to address the above difficulties. We study the problem with the following setting based on the application requirements in surveillance: (1) The clothing of individuals remain unchanged across different scenarios. (2) The individual to be re-identified should be in a moderate resolution, (e.g., $\ge\,120$ pixels in height). Our approach builds a compositional part-based template to represent the target individual and matches the template with input images by employing a stochastic cluster sampling algorithm, as illustrated in Fig.~\ref{fig:framework}.


We organize the template of a query individual with an expressive tree representation that can be produced in a very simple way. We perform the human body part detectors~\cite{PictorRevisit,ViewPictor} on several reference images of the individual, and the images of detected parts are grouped according to their semantics. That is, a human template is decomposed into body parts, e.g., head, torso, arms, each of which associates with a number of part instances. Note that we can prune the instances sharing very similar appearances with others. This expressive template fully exploit information from multiple reference images to capture well appearance variability, partially motivated by the recently proposed hierarchical and part-based models in object recognition~\cite{AOHumParse,AOTObjRec,GeneralSample}. Specifically, several possible instances (namely proposals), extracted from different references, exist at each part in the template, and we regard this representation as the multiple-instance-based compositional template (MICT). As a result, new appearance configurations can be composed by the part proposals in the MICT. One may question the scalability issue for building such a customized template. We argue that the critical concern is accurately identifying the target in realistic scenarios, e.g., searching for one suspect across scenes, rather than processing numbers of targets at the same time.

In the inference stage, the body part detectors are initially utilized to generate possible part locations in the scene shot, and human re-identification is then posed as the task of part-based template matching. Unlike traditional matching problems, the multiple part proposals in the MICT make the search space of matching combinatorially large, as the part proposals need to be activated alone with the matching process. Handling the false alarms and misdetections by the part detectors is also a non-trivial issue during matching. Inspired by recent studies in cluster sampling~\cite{SWCuts,LayerGM,C4}, we propose a stochastic algorithm to solve the compositional template matching.

The matching algorithm is designed based upon the candidacy graph, where each vertex denotes a pair of matching part proposals, and each edge link represents the contextual interaction (i.e. the compatible or the competitive relation) between two matching pairs. Compatible relations encourage vertices to activate together, while competitive relations depress conflicting vertices being activated at the same time. Specifically, two vertices are encouraged to be activated together, as they are kinematically or symmetrically related, whereas two vertices are constrained that only one of them can be activated, as they belong to the same part type or overlap. The algorithm iterates in two steps for optimal matching solution searching. (i) It forms several possible partial matches (clusters) by turning off the edge links probabilistically and deterministically. (ii) It activates clusters to confirm partial matches, leading to a new matching solution that will be accepted by the Markov Chain Monte Carlo (MCMC) mechanism~\cite{SWCuts}. Note that body parts are allowed to be unmatched to cope with occlusions.

The main contributions of this paper are two-fold. First, we propose a novel formulation to solve human re-identification by matching the composite template with cluster sampling. Second, we present a new database including realistic and general challenges for human re-identification, which is more complete than existing related databases.




\vspace{-2mm}
\section{Related Work} \label{sec:Literature}
\vspace{-1mm}

In literature, previous works of human re-identification mainly focus on constructing and selecting distinctive and stable human representation, and they can be roughly divided into the following two categories.

{\bf Global-based methods} define a global appearance human signature with rich image features and match given reference images with the observations~\cite{ELF,ETHZReid,SDALF}. For example, D. Gray et al. propose the feature ensemble to deal with viewpoint invariant recognition. Some methods improve the performance by extracting features with region segmentation~\cite{PrinAxis,SACM,DCDSign}. Recently, advanced learning techniques are employed for more reliable matching metrics~\cite{ReidRDCConfer}, more representative features~\cite{FeatureReid}, and more expressive multi-valued mapping function~\cite{ImplicitTransReid}. Despite acknowledged success, this category of methods often has problems to handle large pose/view variance and occlusions.

{\bf Compositional approaches} re-identify people by using part-based measures. They first localize salient body parts, and then search for part-to-part correspondence between reference samples and observations. These methods show promising results on very challenging scenarios~\cite{iLIDS}, benefiting from powerful part-based object detectors. For example, N. Gheissari et al.~\cite{STAppear} adopt a decomposable triangulated graph to represent person configuration, and the pictorial structures model for human re-identification is introduced~\cite{CPS}. Besides, modeling contextual correlation between body parts is discussed in~\cite{SCR}.

Many works~\cite{STAppear,SDALF,CPS} utilize multiple reference instances for individual, i.e. multi-shot approaches, but they omit occlusions and conjunctions in the target images and re-identify the target by computing a one-to-many distance, while we explicitly handle these problems by exploiting reconfigurable compositions and contextual interactions during inference.


The rest of this paper is organized as follows. We first introduce the representations in Section~\ref{sec:Represent}, and then discuss the inference algorithm in Section~\ref{sec:Algorithm}. The experimental results are shown in Section~\ref{sec:Experiments}, and finally comes the conclusion in Section~\ref{sec:Conclusion}.

\vspace{-2mm}
\section{Representation} \label{sec:Represent}
\vspace{-1mm}

In this section, we first introduce the definition of multiple-instance-based compositional template, and then present the problem formulation of human re-identification.

\vspace{-1mm}
\subsection{Compositional Template} \label{sec:Template}
\vspace{-1mm}

In this work, we present a compositional template to model human with huge variations.

A human body is decomposed into $N=6$ parts: head, torso, upper arms, forearms, thighs and calfs, and each limb is further decomposed into two symmetrical parts (i.e. left and right), as shown in Fig.~\ref{fig:relations}(a)). Each part $g$ is modeled as a rectangle and indicated by a $5$-tuple $(t,x,y,\theta,s)$, where $t$ denotes the part type, $x$ and $y$ the part center coordinates, $\theta$ the part orientation, $s$ the part relative scale, as widely employed in pictorial structures model~\cite{PictorOR,PictorRevisit}. The multiple-instance-based compositional template (MICT) $T$ is defined as
\vspace{-1mm} \begin{equation}
    T = \{\,T_i :\; T_i = \{g\}\;\}_{i=1}^N,
\vspace{-1mm} \end{equation}
where $g$ denotes a part proposal and $T_i$ the set of proposals for the $i$th part in template.




Given reference images of an individual, the MICT is constructed as follows.

We first employ body part detectors to scan every reference image and obtain detection scores for all body parts. The training and detecting process of part detectors closely follows~\cite{ViewPictor}. Given detection scores, we further prune impossible part configurations by several strategies: (i) For all parts, the firing detection is pruned if the overlap rate of foreground mask (done by background subtraction) is less than $75\%$. (ii) The reference image is segmented into $4$ horizonal strips with equal height. Head is detected in the first strip (the first to fourth top to bottom), parts of upper body (i.e. torso, upper arms and forearms) in the second, and parts of lower body (i.e. thighs and calfs) the rest. Finally, we apply non-maximum suppression and collect the $K$ proposals with highest responses for each part from all reference images.


Given target images (scene shots) to be matched, we can obtain the target proposal set $G$ by a similar process as constructing the MICT, except the firing detection being pruned only by the foreground mask. Considering realistic complexities in surveillance, there probably exist large numbers of detection false alarms in the target proposal set $G$.

%

%
%




\vspace{-1mm}
\subsection{Candidacy Graph} \label{sec:CandiGraph}
\vspace{-1mm}

Given the template $T$ and the target proposal set $G$, the problem of human re-identification can be posed as the task of part-based template matching and solved by two steps: (i) activating one proposal for each part in $T$, (ii) finding the match in $G$.


We define the set of activated part proposals $\Psi$ from $T$, each of which corresponds to a certain part:
\vspace{-1mm} \begin{equation}
    \Psi \,=\, \{\,\Psi_i:\; l(\Psi_i) = 1,\;\; \Psi_i \in T_i\,\}_{i=1}^{N}.
\vspace{-1mm} \end{equation}
The binary label $l(\cdot)$ indicates whether the proposal is activated or remains inactivated, i.e. $l(\cdot) = 1$ for activated and $l(\cdot) = 0$ for inactivated. The set of matched part proposals from $G$ can be defined as
\vspace{-1mm} \begin{equation}
    \Phi \,=\, \{\,\Phi_i:\; l(\Phi_i) = 1,\;\; \Phi_i \in G_i \cup \{\emptyset\}\,\}_{i=1}^N,
\vspace{-1mm} \end{equation}
where $\Phi_i$ maps the activated proposal of the $i$th part in $T$ to a proposal in $G$. Note that $\Psi_i$ not necessarily has a match $\Phi_i$ (i.e. $l(\emptyset)=1$), in case the matched part is occluded or missed in $G$.

To solve these two steps simultaneously, we propose a candidacy graph representation and further formulate the problem by graph labeling. We define the candidacy graph $\mathbb{G} = \slangle \mathbb{C},\mathbb{E} \srangle$, where each vertex $c_i \in \mathbb{C}$ denotes a candidate matching pair $(\Psi_i,\Phi_i)$. A similar binary label $l(c_i)$ is employed to indicate whether a matching pair $c_i$ is activated or not. Solving the matching problem is equivalent to labeling vertices $\mathbb{C}$ in the candidacy graph $\mathbb{G}$. The label set $L$ is thus defined as
\vspace{-1mm} \begin{equation}
    L \,=\,\{l(c_i) = l_i: l_i \in \{0,1\}, i=1,...,|\mathbb{C}|, c_i \in \mathbb{C}\}.
\vspace{-1mm} \end{equation}

Each edge $e_{ij} = \slangle c_i,c_j \srangle$ in $\mathbb{G}$ denotes the relation between two matching pairs $c_i$ and $c_j$. We incorporate two kinds of relations, i.e. compatible and competitive relations, to model the contextual interactions in scene shots. In the following discussion, we drop the notation of edge index $ij$ for notation simplicity.



\begin{figure}[ptb]
\begin{center}
\includegraphics[width=3.3in]{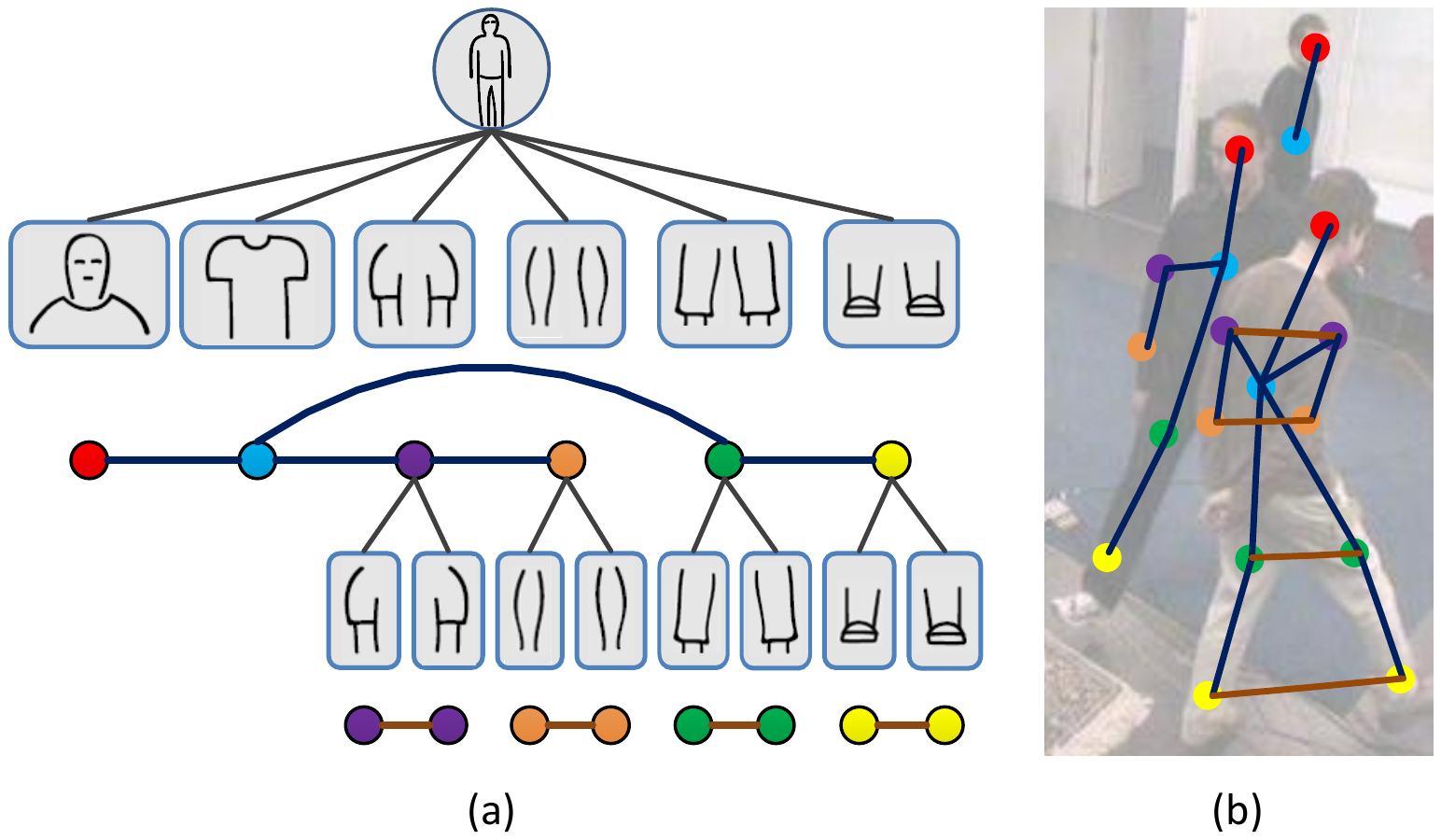}
\end{center}
\vspace{-4mm}
\caption{An illustration of compatible relations. (a) Kinematics (navy blue edges) and symmetry (brown edges) relations within the compositional template. (b) An example to show how target part proposals are coupled together by kinematics and symmetry relations.}
\label{fig:relations}
\vspace{-3mm}
\end{figure}

\begin{figure*}[ptb]
\begin{center}
\includegraphics[width=6in]{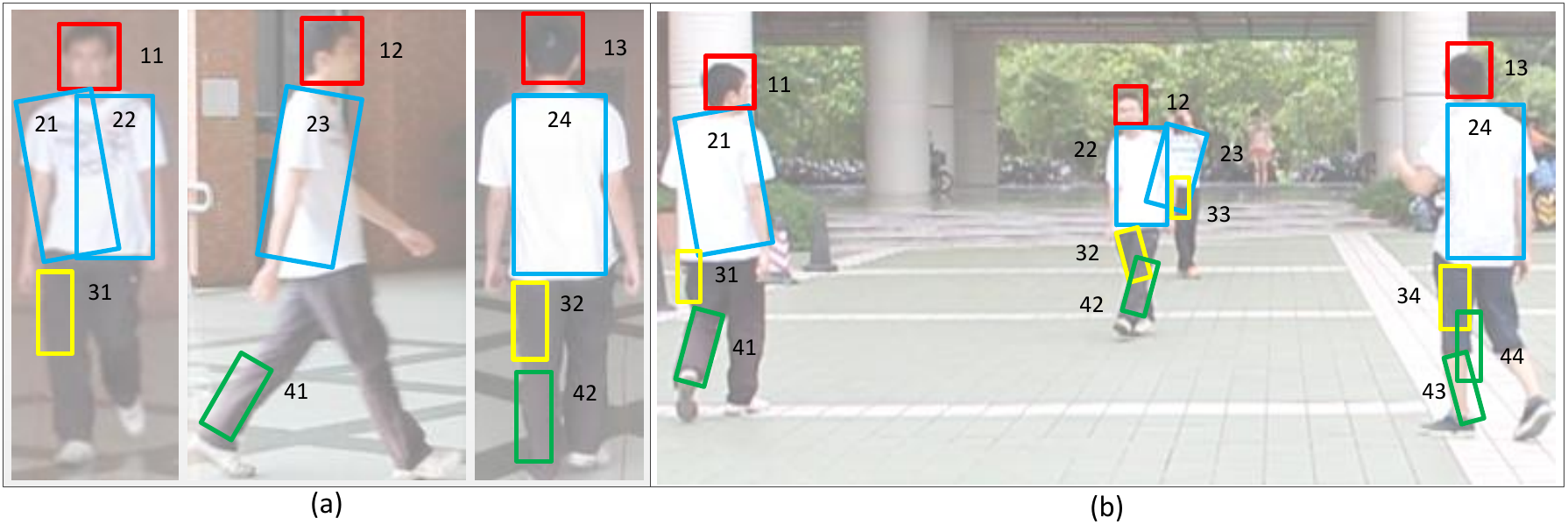}
\end{center}
\vspace{-5mm}
\caption{A re-identification example used to illustrate our inference algorithm. Given (a) reference images and (b) a scene shot, proposals of four parts: head, torso, left thigh, left calf are drawn and numbered in the image. Note that we omit the other parts and only keep few proposals for clear specification.}
\label{fig:example}
\vspace{-4mm}
\end{figure*}

{\bf Compatible relations} encourage matching pairs to activate together in matching. We represent compatible relations as how two target part proposals are coupled together and mainly explore two cases: (i) kinematics relations for coupling kinematic dependent parts. (ii) symmetry relations for coupling symmetrical parts. That is,
\vspace{-1mm} \begin{equation} \label{equ:posedge}
    p_e^+(c_i,c_j) \,=\, \begin{cases}
        & p_k(\Phi_i,\Phi_j),\;\quad \slangle t_i,t_j \srangle \in \text{Kin},\\
        & p_s(\Phi_i,\Phi_j),\;\quad \slangle t_i,t_j \srangle \in \text{Sym},
    \end{cases}
\vspace{-1mm} \end{equation}
where $t_i$ and $t_j$ denotes the part type of $\Phi_i$ and $\Phi_j$, respectively.

(i) Kinematics relations describe spatial relationship between kinematic dependent parts (navy blue edges in Fig.~\ref{fig:relations}(a)). The spatial distribution $p_k(\Phi_i,\Phi_j)$ between two proposals $\Phi_i$ and $\Phi_j$ is modeled as a zero-mean Gaussian distribution under the coordinate system of their connected joint:
\vspace{-2mm} \begin{equation}
    p_k(\Phi_i,\Phi_j) \,\varpropto\, N(F_{ji}(\Phi_i)-F_{ij}(\Phi_j),0,\Sigma_{ji}),
\vspace{-1mm} \end{equation}
where $F_{ji}(\cdot)$ and $F_{ij}(\cdot)$ are the transformations of $\Phi_i$ and $\Phi_j$ from image coordinate system to joint coordinate system. For detailed explanations, see~\cite{PictorOR,PictorRevisit}.

In the experiment, kinematics relations are learnt from reference images with body part annotations.

(ii) Symmetry relations measure the appearance similarity between symmetrical parts (brown edges in Fig.~\ref{fig:relations}(a)). We suppose symmetrical parts from the same individual tend to share similar appearance while those from different individuals don't. Therefore the symmetry relations are represented as
\vspace{-2mm} \begin{equation}
    p_s(\Phi_i,\Phi_j) \,\varpropto\, e^{-D(\Phi_i,\Phi_j)},
\vspace{-1mm} \end{equation}
where $D(\cdot)$ measures the distance between two part proposals and is defined in Equ.(\ref{equ:distance}).

We give an example to illustrate how kinematics relations and symmetry relations work in scene shots, as shown in Fig.~\ref{fig:relations}(b). Note that we omit certain part proposals for clear specification.

{\bf Competitive relations} depress conflicting matching pairs being activated at the same time. We also develop two cases for competitive relations: (i) Two target proposals with the same part type cannot be activated simultaneously. (ii) The overlapped region between two target part proposals should only be compared once. That is,
\vspace{-1mm} \begin{equation} \label{equ:negedge}
    p_e^-(c_i,c_j) \varpropto \begin{cases}
        & \quad\quad\, 1,  \quad\quad\quad t_i=t_j,\\
        & e^{\lambda \cdot \mathbf{IoU}(\Phi_i,\Phi_j) },\; t_i\ne t_j, \Phi_i \cap \Phi_j \ne \emptyset,
    \end{cases}
\vspace{-1mm} \end{equation}
where $\mathbf{IoU}(\Phi_i,\Phi_j)$ indicates the overlap intersection-over-union between $\Phi_i$ and $\Phi_j$, $\lambda$ is a scaling constant.

%

\begin{figure}[ptb]
\vspace{-3mm}
\begin{center}
\includegraphics[width=3.3in]{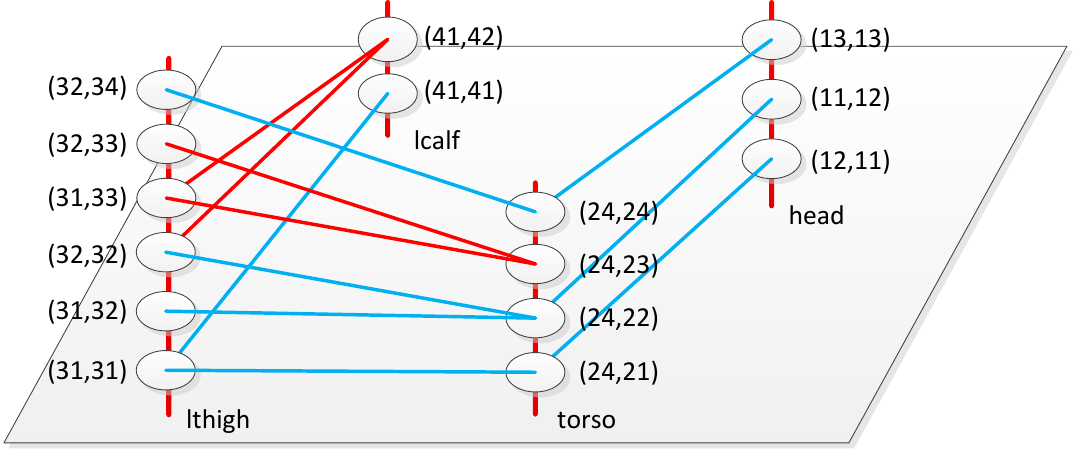}
\end{center}
\vspace{-4mm}
\caption{An illustration of the candidacy graph representation. We visualize the candidacy graph of Fig.~\ref{fig:example}. In the graph, vertices denote candidate matches, blue and red edges indicate compatible and competitive edges between vertices, respectively.}
\vspace{-3mm}
\label{fig:candigraph}
\end{figure}

An illustration of the candidacy graph representation is shown in Fig.~\ref{fig:candigraph}, corresponding to the example in Fig.~\ref{fig:example}.

\begin{figure*}[ptb]
\begin{center}
\includegraphics[width=6.4in]{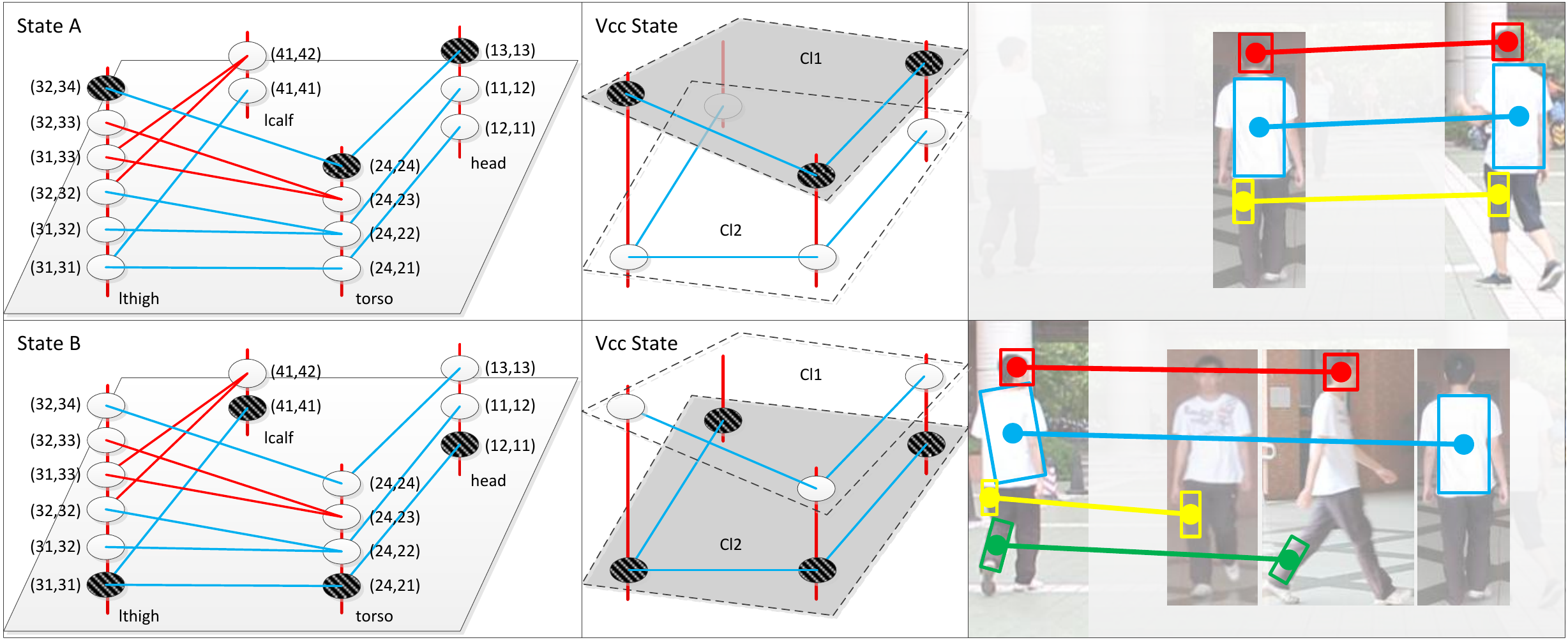}
\end{center}
\vspace{-4mm}
\caption{An illustration of one transition in composite cluster sampling. The first row and the second row denote labels of part proposals, labels of the composite cluster and matching configurations of two successive states (A and B) in one reversible transition, respectively.}
\label{fig:clustersample}
\vspace{-4mm}
\end{figure*}

In summary, the problem of matching the template $T$ to the target proposal set $G$ can be represented as
\vspace{-1mm} \begin{equation}\label{equ:model}
    M \,=\,(N_u,N_s,L),
\vspace{-1mm} \end{equation}
where $N_u$ denotes the number of unmatched part pairs and $N_s$ the number of scales of the activated proposals, and they can be computed from the labeling set $L$. According to Bayes' Rule, $M$ can be solved by maximizing a posterior probability:
\vspace{-1mm} \begin{equation} \label{equ:posterior} \begin{aligned}
    M^* &\,=\,\underset{M}{\arg\,\max}\;p(M|T,G)\\
        &\,=\,\underset{M}{\arg\,\max}\;p(M|\mathbb{G})\\
        &\,\varpropto\,\underset{M}{\arg\,\max}\;p(\mathbb{G}|M) \cdot p(M).
\end{aligned} \vspace{-1mm} \end{equation}

{\bf Likelihood} $p(\mathbb{G}|M)$ measures the appearance similarity between the template and the matching target. Assuming the appearance similarity of each matching pair is independent, then $p(\mathbb{G}|M)$ can be factorized into
\vspace{-1mm} \begin{equation}
    p(\mathbb{G}|M) \,\varpropto\, \prod_{c_i \in \mathbb{C}}\, p(c_i|l_i) \,=\, \prod_{ l_i = 1, c_i \in \mathbb{C}}\, e^{-D(\Psi_i,\Phi_i)}\,,
\vspace{-1mm} \end{equation}
where $D(\cdot)$ denotes the distance between two proposals.


We adopt modified HSV color histogram~\cite{CPS} and MSCR descriptor~\cite{MSCR} to describe the visual statistics for each part proposal, which has been widely used in existing human re-identification studies~\cite{SDALF,CPS}. The distance $D(\cdot)$ between two arbitrary proposals $g_i$ and $g_j$ is defined as
\vspace{-1mm} \begin{equation} \label{equ:distance}
    D(g_i,g_j) = D_{Bh}(H_i,H_j) + D_{\text{MSCR}}(Dc_i,Dc_j),
\vspace{-1mm} \end{equation}
where $H$ denotes the normalized HSV color histogram, $Dc$ the MSCR descriptor, $D_{Bh}(\cdot)$ and $D_{\text{MSCR}}(\cdot)$ the Bhattacharyya distance and the distance defined in~\cite{SDALF}, respectively.

{\bf Prior} $p(M)$ penalizes the undesired activation of matching pairs (e.g. missing parts) and matching inconsistency among the activated matching pairs. We define $p(M)$ as
\vspace{-1mm} \begin{equation} \begin{aligned}
    p(M) \,&=\, p(N_u) \cdot p(N_s) \cdot p(L) \\
         \,&\varpropto\, \exp\{-\alpha_u N_u-\alpha_s N_s\}\cdot p(L),
\end{aligned} \vspace{-1mm} \end{equation}
where $\alpha_u$ and $\alpha_s$ are corresponding parameters for $N_u$ and $N_s$, respectively.

$p(L)$ imposes constraints on the edge links among activated vertices, that is
\vspace{-1mm} \begin{equation}
    p(L) \varpropto \underset{l_i=l_j=1, e \in \mathbb{E}^+}{\prod} p_e^+(c_i,c_j) \underset{l_i=l_j=1, e \in \mathbb{E}^-}{\prod} (1 \sminus p_e^-(c_i,c_j)),
\vspace{-1mm} \end{equation}
where $\mathbb{E}^+$ and $\mathbb{E}^-$ indicate the compatible edges and competitive edges in the candidacy graph $\mathbb{G}$, respectively.

\vspace{-2mm}
\section{Inference Algorithm} \label{sec:Algorithm}
\vspace{-1mm}

In a scene shot containing multiple individuals, matching the template to the target becomes an extremely complicated problem. For example, in Fig.~\ref{fig:example}, the four individuals in the shot all share similar appearance with the template. As a result, solving Equ.(\ref{equ:posterior}) probably leads to a local optimal solution. In this case, popular inference algorithms, such as EM, Belief Propagation and Dynamic Programming, are easily struck and thus fail to re-identify the correct target (i.e. finding global optimal solution), while Composite Cluster Sampling, as introduced in~\cite{LayerGM,C4}, overcomes this problem by jumping from partial coupling matches in each MCMC step. Therefore, we employ Composite Cluster Sampling to search for optimal match between the template and the correct target.

Composite Cluster Sampling algorithm consists of the following two steps:

(I) {\bf Generating a composite cluster}. Given a candidacy graph $\mathbb{G} = \slangle \mathbb{C},\mathbb{E} \srangle$ and the current matching state $M$, we first separate graph edges $\mathbb{E}$ into two sets: set of inconsistent edges $\{e \in \mathbb{E}^+: l(c_i) \ne l(c_j)\} \cup \{e \in \mathbb{E}^-: l(c_i) = l(c_j)\}$ (i.e. edges violating current state) and set of consistent edges in the other two cases. Next we introduce a boolean variable $\omega_e \in \{1,0\}$ to indicate an edge is being turned on or turned off. We turn off inconsistent edges deterministically and turn on every consistent edge with its edge probability $p_e$. Afterwards, we regard candidates connected by "on" positive edges as a cluster $Cl$ and collect clusters connected by "on" negative edges to generate a composite cluster $V_{cc}$.

(II) {\bf Relabeling the composite cluster}. In this step, we randomly choose a cluster from the obtained composite cluster $V_{cc}$ and flip the labels of the selected cluster and its conflicting clusters (i.e. the clusters connected with the selected cluster), which generates a new state $M'$. To find a better state and achieve a reversible transition between two states $M$ and $M'$, the acceptance rate of the transition from state $M$ to state $M'$ is defined by a Metropolis-Hastings method~\cite{MHAlgorithm}:
\vspace{-2mm} \begin{equation}
    \alpha(M \sto M') = \min(1,\frac{q(M' \sto M) \cdot p(M'|\mathbb{G})}{q(M \sto M') \cdot p(M|\mathbb{G})}),
\vspace{-1mm} \end{equation}
where $q(M' \sto M)$ and $q(M \sto M')$ denote the state transition probability, $p(M'|\mathbb{G})$ and $p(M|\mathbb{G})$ the posterior defined in Equ.(\ref{equ:posterior}).

Following instructions in~\cite{SWCuts}, the state transition probability ratio is computed by
\vspace{-1mm} \begin{equation} \label{equ:transrate}\begin{aligned}
    \frac{q(M' \sto M)}{q(M \sto M')} &\varpropto \frac{q(V_{cc}|M')}{q(V_{cc}|M)} \\
                                    &\varpropto \frac{\prod_{e \in \mathcal{E}^+_{M'}}(1-\rho_e^+)\prod_{e \in \mathcal{E}^-_{M'}}(1-\rho_e^-)} {\prod_{e \in \mathcal{E}^+_M}(1-\rho_e^+)\prod_{e \in \mathcal{E}^-_M}(1-\rho_e^-)},
\end{aligned} \vspace{-1mm} \end{equation}
where $\mathcal{E}^+$ and $\mathcal{E}^-$ denote the sets of positive and negative edges being turned off around $V_{cc}$, respectively, that is,
\begin{equation} \begin{aligned}
    & \mathcal{E}^+ = \{e \in\mathbb{E}^+:c_i\in V_{cc}, c_j \not\in V_{cc}, l(c_i)=l(c_j)\}, \\
    & \mathcal{E}^- = \{e \in\mathbb{E}^-:c_i\in V_{cc}, c_j\not\in V_{cc}, l(c_i)\ne l(c_j)\}.
\end{aligned} \end{equation}
Note that the subscript of $\mathcal{E}^+$, $\mathcal{E}^-$ in Equ.(\ref{equ:transrate}) indicates the current state and is omitted for simplicity in the above definition.

We show an example of one transition in composite cluster sampling in Fig.~\ref{fig:clustersample}. In this figure, $V_{cc}$ contains two clusters $\{Cl_1,Cl_2\}$. In state A, $Cl_1$ is activated and the conflicting cluster $Cl_2$ is deactivated while in state B labels of $Cl_1$ and $Cl_2$ are flipped. The transition from state $A$ to state $B$ achieves a fast jump between two kinds of partial coupling matches and coincides with an individual-to-individual comparison in re-identification.

Applying the above mechanism, we summarize the inference algorithm in Algorithm 1.

\begin{figure}[htb]
\vspace{-2mm}
\begin{center}
\includegraphics[width=3.2in]{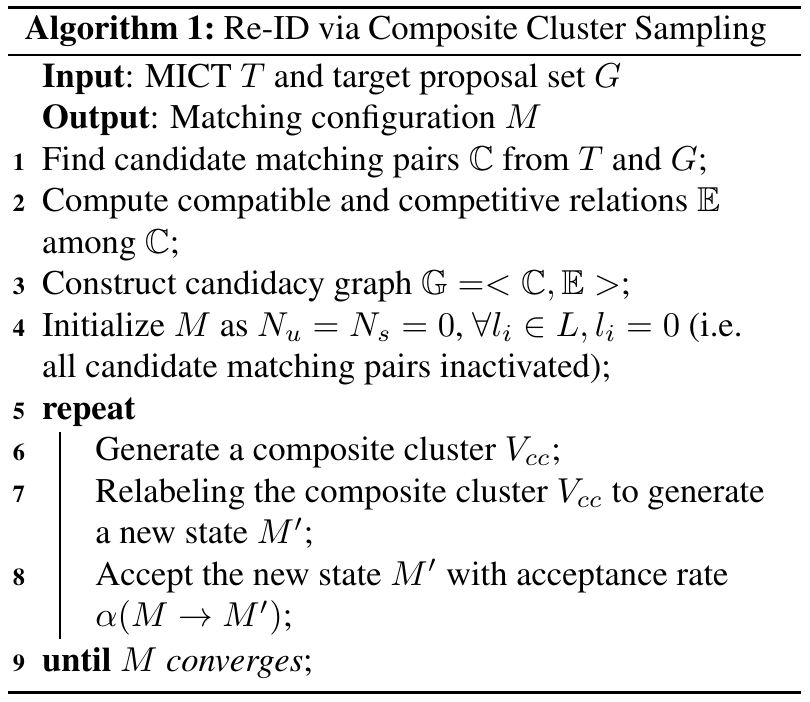}
\end{center}
\label{fig:algorithm}
\vspace{-6mm}
\end{figure}

%
%

\vspace{-2mm}
\section{Experiments} \label{sec:Experiments}
\vspace{-1mm}

In this section, we first introduce the datasets and the parameter settings, and then show our experimental results as well as component analysis of the proposed approach.

\vspace{-1mm}
\subsection{Datasets and Settings} \label{sec:Dataset}
\vspace{-1mm}

We validate our method on three public databases as follows.

(i) {\bf VIPeR dataset}\footnote{Available at \url{www.umiacs.umd.edu/~schwartz/datasets.html}}. It is commonly used for human re-identification, containing $632$ people in outdoor, and there are $2$ images for each individual.

(ii) {\bf EPFL dataset}\footnote{Available at \url{cvlab.epfl.ch/data/pom/}}. This database is very challenging, originally proposed for tracking in multi-views~\cite{EPFL}. It consists of $5$ different scenarios that are filmed by three or four cameras from different angles. For evaluating our method, we extract individuals from the original videos and annotate each of them with ID and location (bounding box). In total, there are $70$ reference images for $30$ different individuals, (normalized to $175$ pixels in height), and $80$ shots in $360 \times 288$, which contain $294$ targets to be re-identified.

(iii) {\bf CAMPUS-Human dataset}\footnote{Available at \url{http://vision.sysu.edu.cn/projects/human-reid/}}. We construct this database including general and realistic challenges for people re-identification in surveillance. There are $370$ reference images normalized to $175$ pixels in height, for $74$ individuals, with IDs and locations provided. We present $214$ shots containing $1519$ targets for evaluating methods, and the targets often appear with diverse poses/views, conjunctions and occlusions, see Fig.~\ref{fig:resulttable} (bottom row). Note all images in both EPFL dataset and CAMPUS-Human dataset are captured from the original videos with large time gap to guarantee appearance varieties (unlike ETHZ dataset~\cite{ETHZReid}).


{\em Experiment settings}. For VIPeR dataset, we adopt the common setting that running the algorithm on random partitions containing $316$ pairs. For EPFL and CAMPUS-Human dataset, we randomly select reference images for each individual, and all target images are tested to match. The results on all three datasets are computed by taking average over ten runs. Our approach is evaluated under cases of both single reference image (single-shot, SvsS) and multiple reference images (multi-shot, MvsS, $M = 2,3$).


All the parameters are fixed in the experiments, including $\lambda = 10$ for scaling the overlap $\mathbf{IoU}$, $\alpha_u = 12$ and $\alpha_s = 3$ for penalizing the activation of vertices. We construct the MICT for each individual with their selected reference images. In the re-identification, a number $K$ of body part proposals are generated. In practice, we set $K$ approximately $3$ times the number of individuals in the shot.

We implement our approach with C++ and run the program on a PC with I5 2.8GHZ CPU and 4GB memory. On average, the inference algorithm converges after around $500$ samplings, which costs $2s \sim 40s$. The time cost is related with the complexity of the candidacy graph.

\begin{figure*}[ptb]
\begin{center}
\includegraphics[width=1.69in]{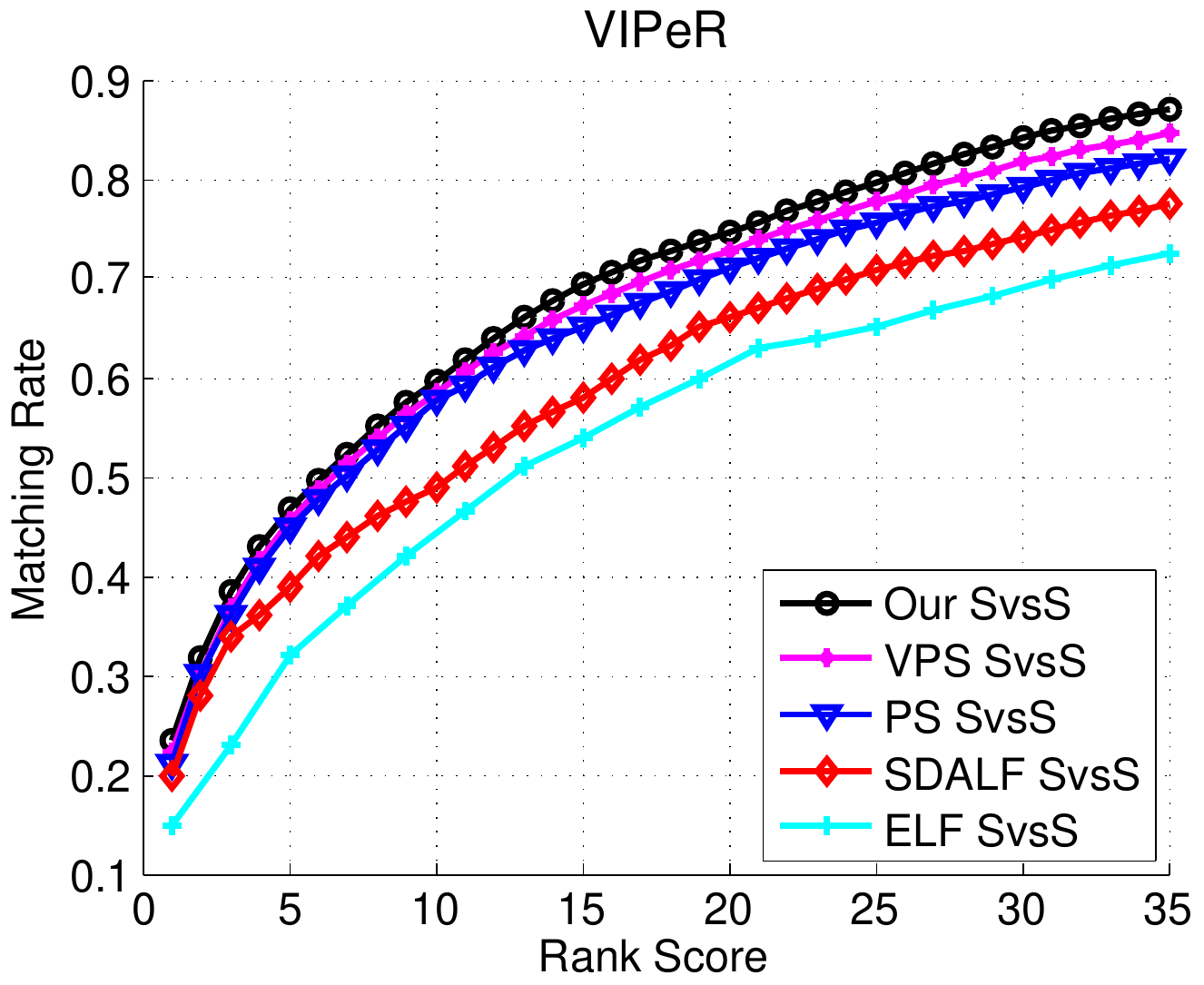}
\includegraphics[width=1.69in]{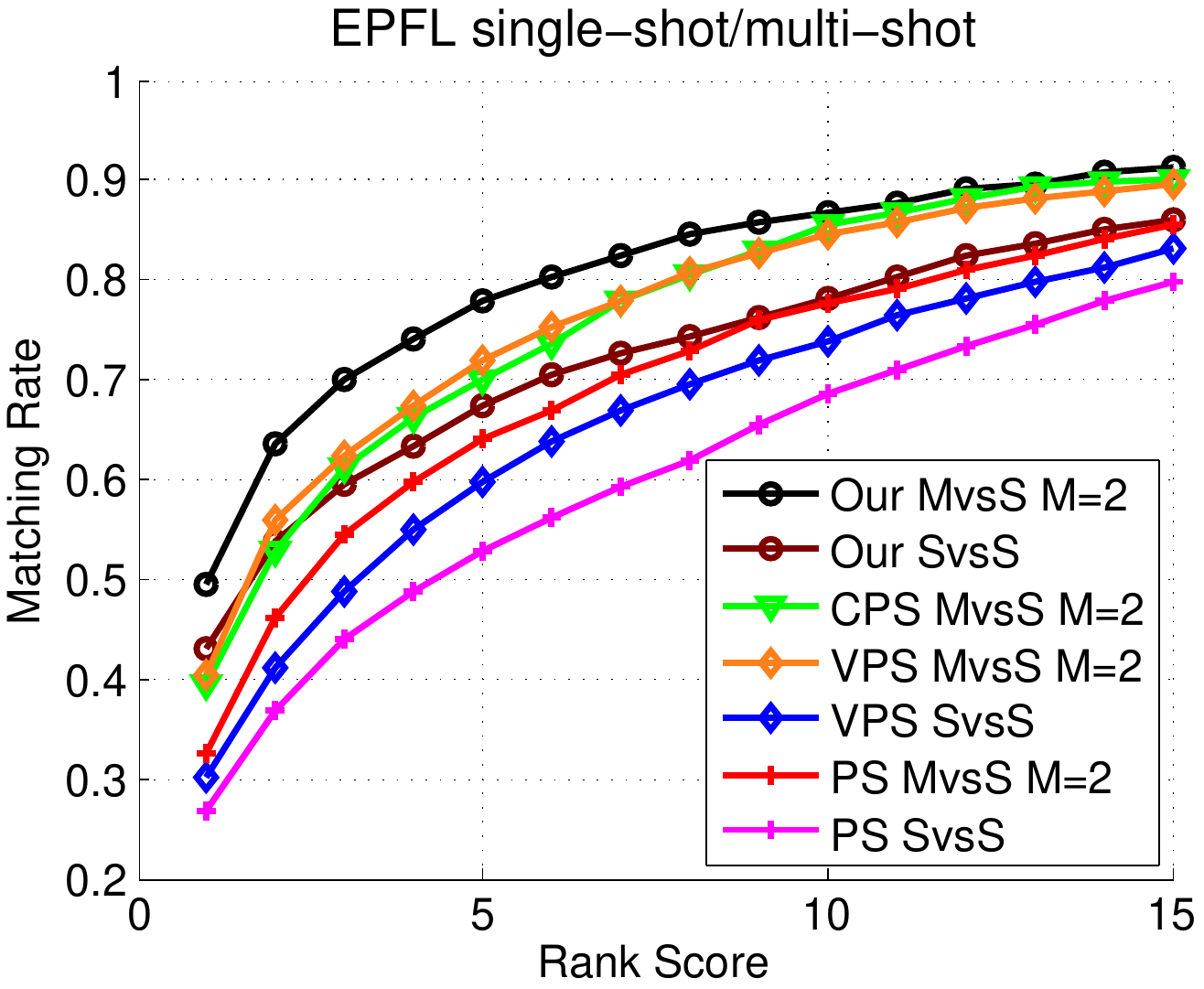}
\includegraphics[width=1.69in]{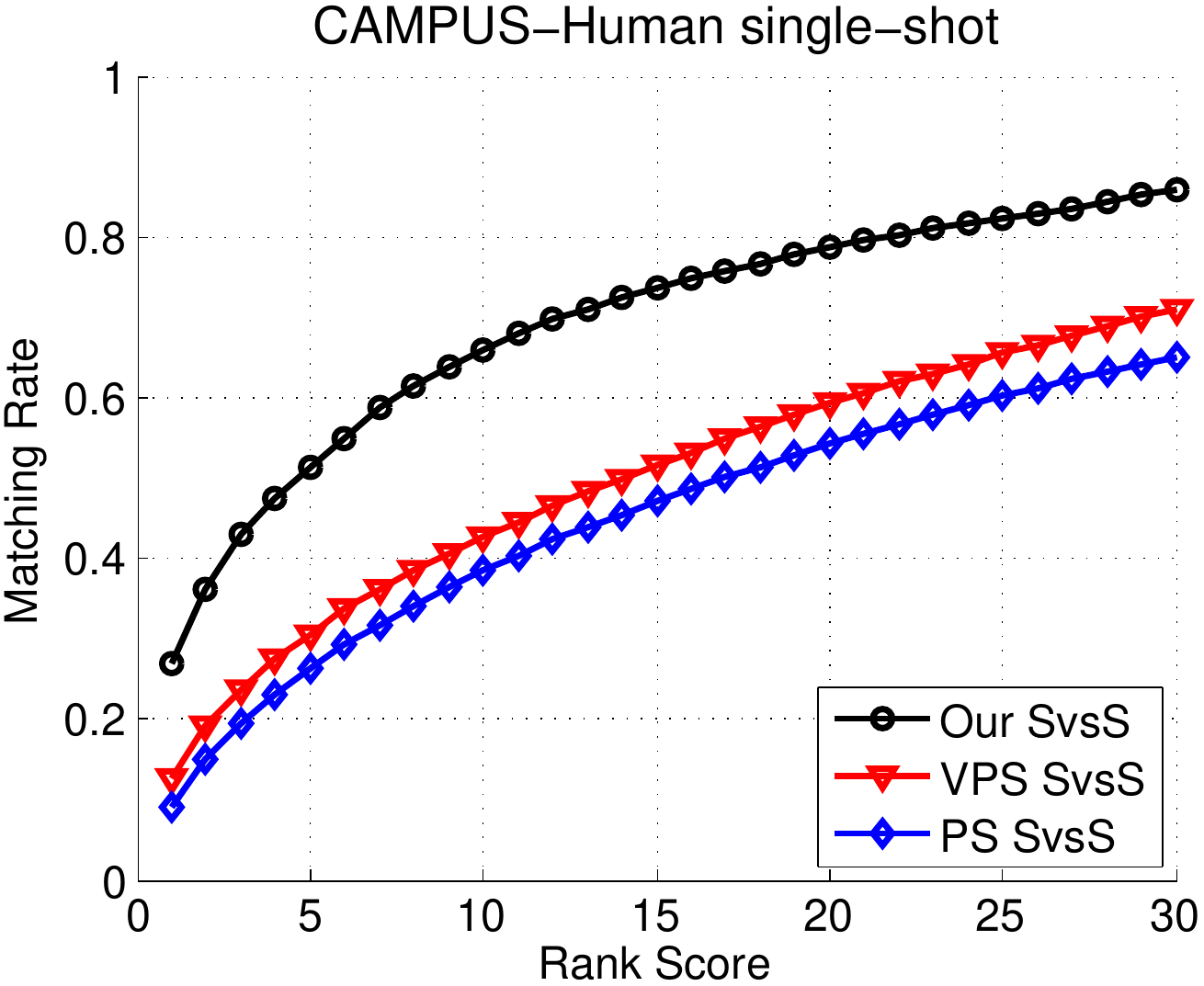}
\includegraphics[width=1.69in]{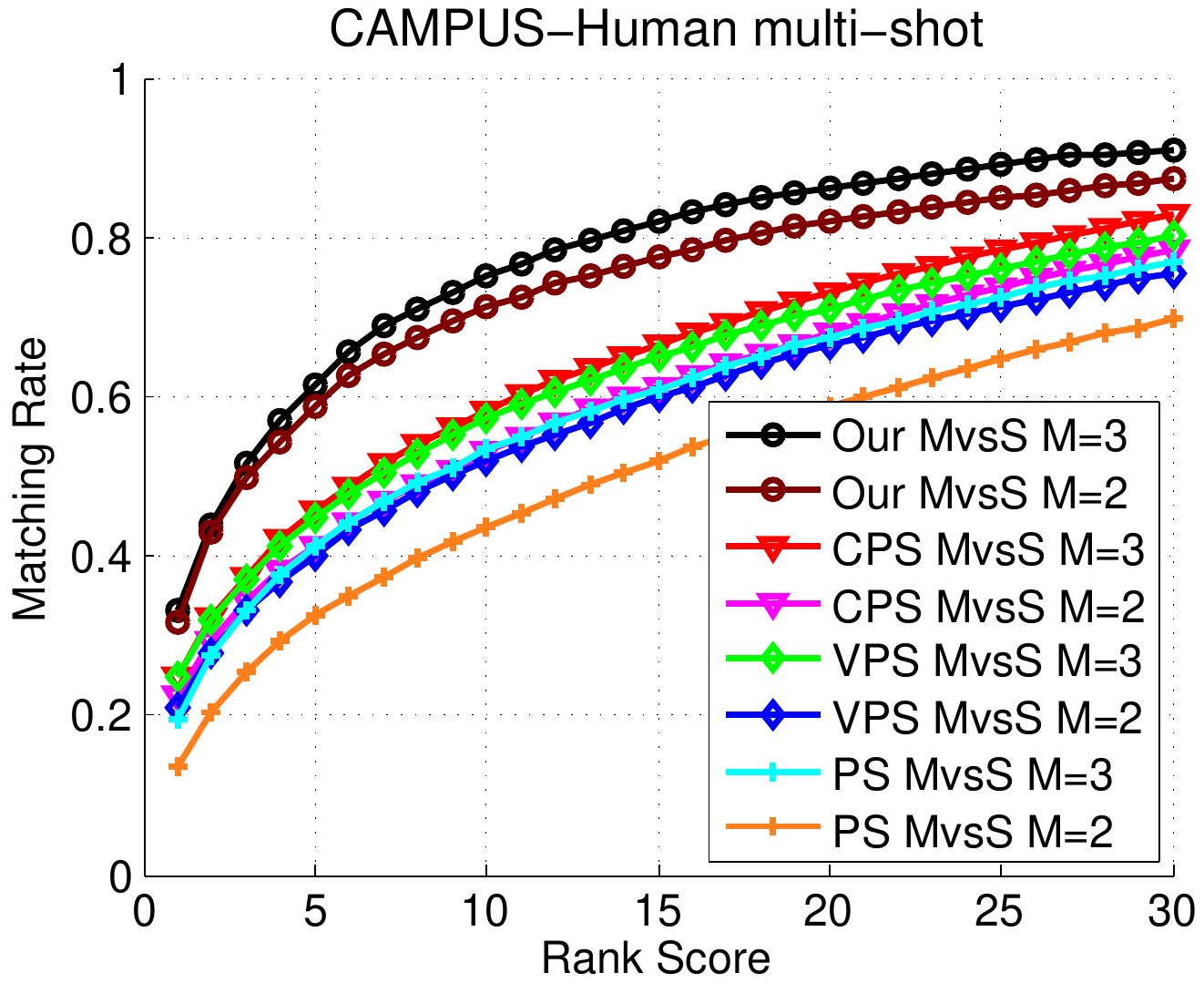}
\end{center}
\vspace{-4mm}
\caption{Performance comparisons using the CMC curves on VIPeR (left1), EPFL (left2), CAMPUS-Human (right1 and right2) datasets. EPFL and CAMPUS-Human datasets are evaluated in both single-shot and multi-shot cases.}
\label{fig:CMCCurves}
\vspace{-4mm}
\end{figure*}

\vspace{-1mm}
\subsection{Experimental Results} \label{sec:ExpResults}
\vspace{-1mm}


%
%
%


We compare our approach with the state-of-the-arts methods: Pictorial Structures (PS)~\cite{PictorRevisit}, View-based Pictorial Structures (VPS)~\cite{ViewPictor}, Custom Pictorial Structures (CPS)~\cite{CPS}, Symmetry-driven Accumulation of Local Features (SDALF)~\cite{SDALF} and Ensemble of Localized Features (ELF)~\cite{ELF}. We adopt the provided code of PS and implement VPS and CPS according to their descriptions. For fair comparison, the same likelihood is employed for PS, VPS and CPS as the proposed method. The results are evaluated by two ways: (i) re-identifying individuals in segmented images, i.e. targets already localized, and (ii) re-identifying individuals from scene shots without provided segmentations.

For the first evaluation, we adopt the cumulative match characteristic (CMC) curve for quantitative analysis, as in previous works~\cite{VIPeRReid,ETHZReid}. The curve reflects the overall ranked matching rates; precisely, a rank $r$ matching rate indicates the percentage of correct matches found in top $r$ ranks. As Fig.~\ref{fig:CMCCurves} shows, we demonstrate the superior performance over the competing approaches in both single-shot case and multi-shot case. And our method yields the best rank $1$ matching rate on EPFL and CAMPUS-Human datasets. We observe that the performance of re-identification can be improved significantly by fully exploiting reconfigurable compositions and contextual interactions in inference. Our performance only improves slightly on VIPeR dataset, as most erroneous matchings are due to severe illumination changes, which has been approved in~\cite{SDALF}.



\begin{table}[pb]
  \vspace{-5mm}
  \centering
  \caption{Matching rate of re-identifying targets in scene shots without provided segmentations.}
    \begin{tabular}{ccc}
    \addlinespace
    \toprule
    Dataset & EPFL  & CAMPUS-Human \\
    \midrule
    \multicolumn{1}{l}{Our M=2}     & {\bf 57/294} & {\bf 215/1519}\\
    \multicolumn{1}{l}{VPS M=2}     & 54/294       & 175/1519\\
    \multicolumn{1}{l}{PS M=2}      & 32/294       & 141/1519 \\
    \midrule
    \multicolumn{1}{l}{Our Single}  & {\bf 50/294} & {\bf 173/1519}\\
    \multicolumn{1}{l}{VPS Single}  & 49/294       & 139/1519\\
    \multicolumn{1}{l}{PS Single}   & 24/294       & 118/1519 \\
    \bottomrule
    \end{tabular}%
  \label{tab:batchresult}%
\end{table}%

\begin{figure*}[ptb]
\begin{center}
\includegraphics[width=\textwidth]{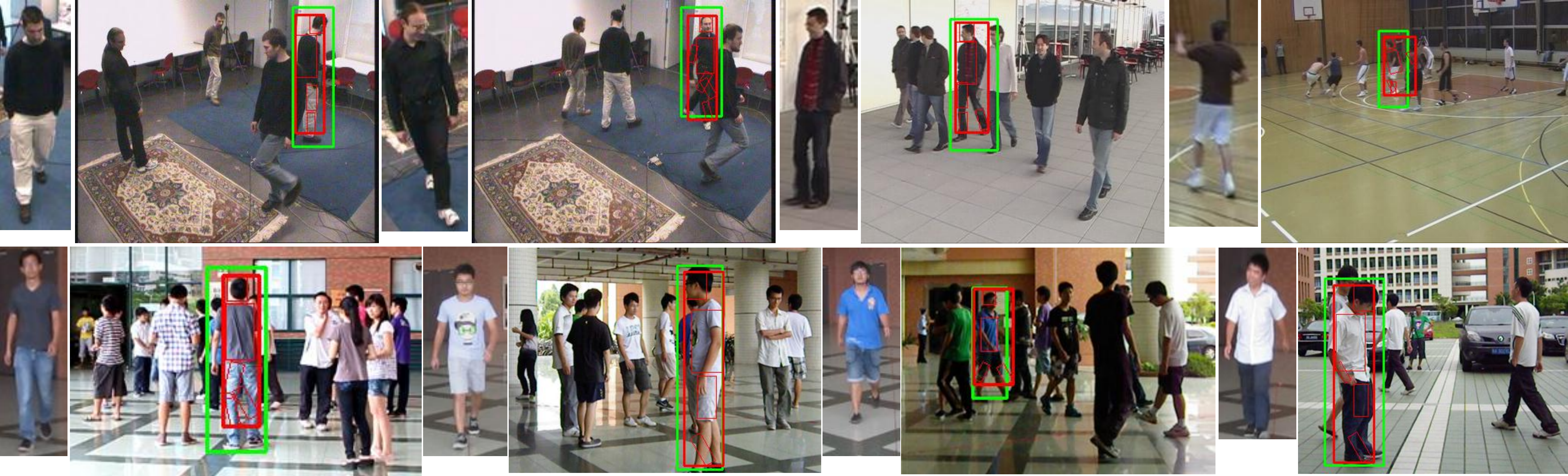}
\end{center}
\vspace{-4mm}
\caption{Results generated by our approach on EPFL and CAMPUS-Human datasets. In each result, the query individual is specified by the image beside the shot. Green boundings denote the target groundtruth location, while red boundings are generated by algorithm.}
\label{fig:resulttable}
\vspace{-4mm}
\end{figure*}

The second test is stricter, since the algorithms should also localize the target during re-identification. We adopt the PASCAL Challenge criterion to evaluate the localization results: a match is counted as the correct match only if the intersection-over-union ratio ($\mathbf{IoU}$) with the groundtruth bounding box is greater than $50\%$. We compare our method with PS~\cite{PictorRevisit}, VPS~\cite{ViewPictor}, which can localize the body at the same time as localizing the parts. The quantitative results are reported in Table~\ref{tab:batchresult}. A number of representative results generated by our method are exhibited in Fig.~\ref{fig:resulttable}. From the results, existing methods perform poor when individuals are not well segmented and scaled to uniform size. In contrast, our method can re-identify challenging target individuals by searching and matching their salient parts and thus achieves better performance. Note the performance of our approach also drops significantly due to inaccurate part localizations and interference of other individuals.




\begin{figure}[htb]
\begin{center}
\includegraphics[width=1.62in]{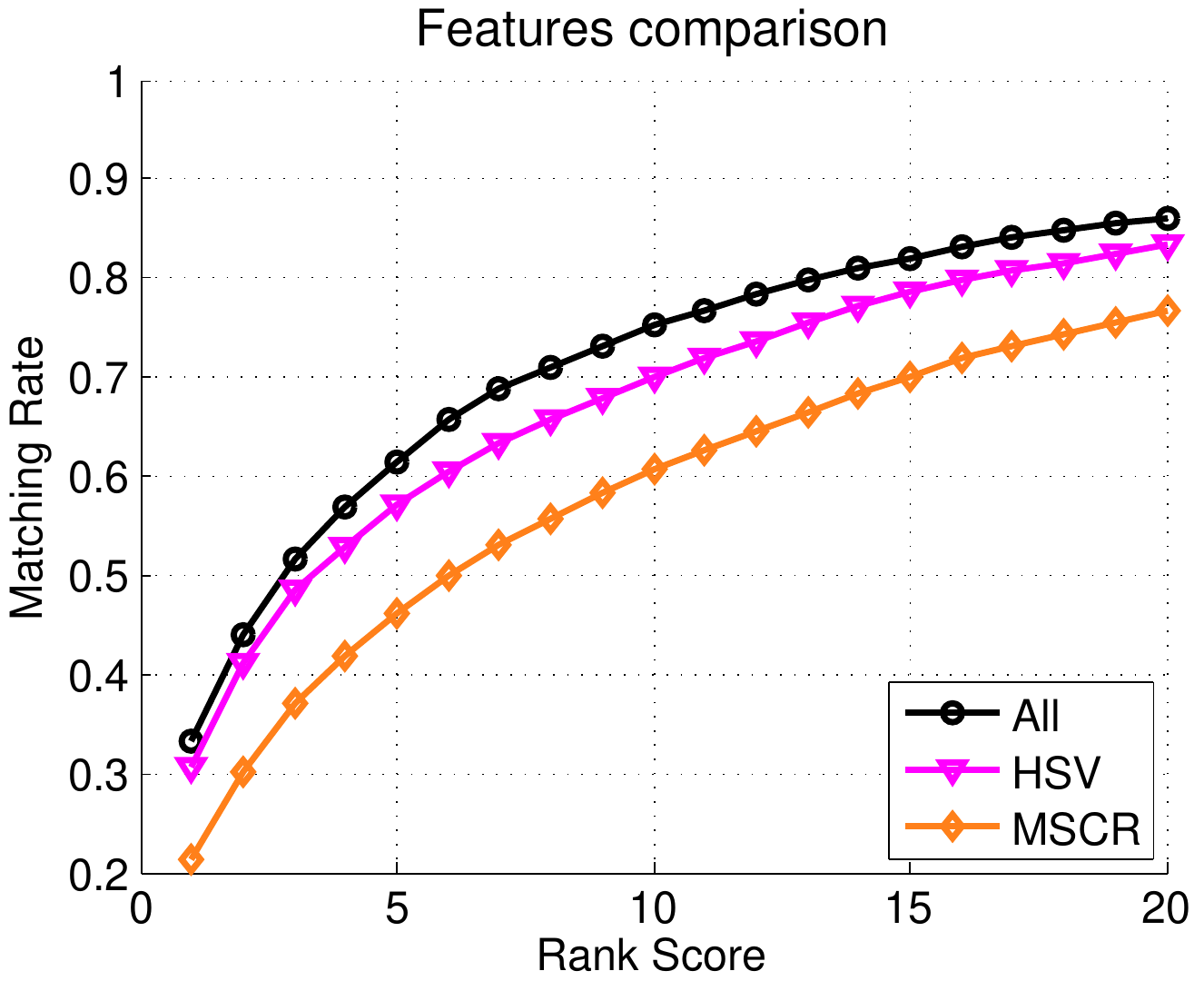}
\includegraphics[width=1.62in]{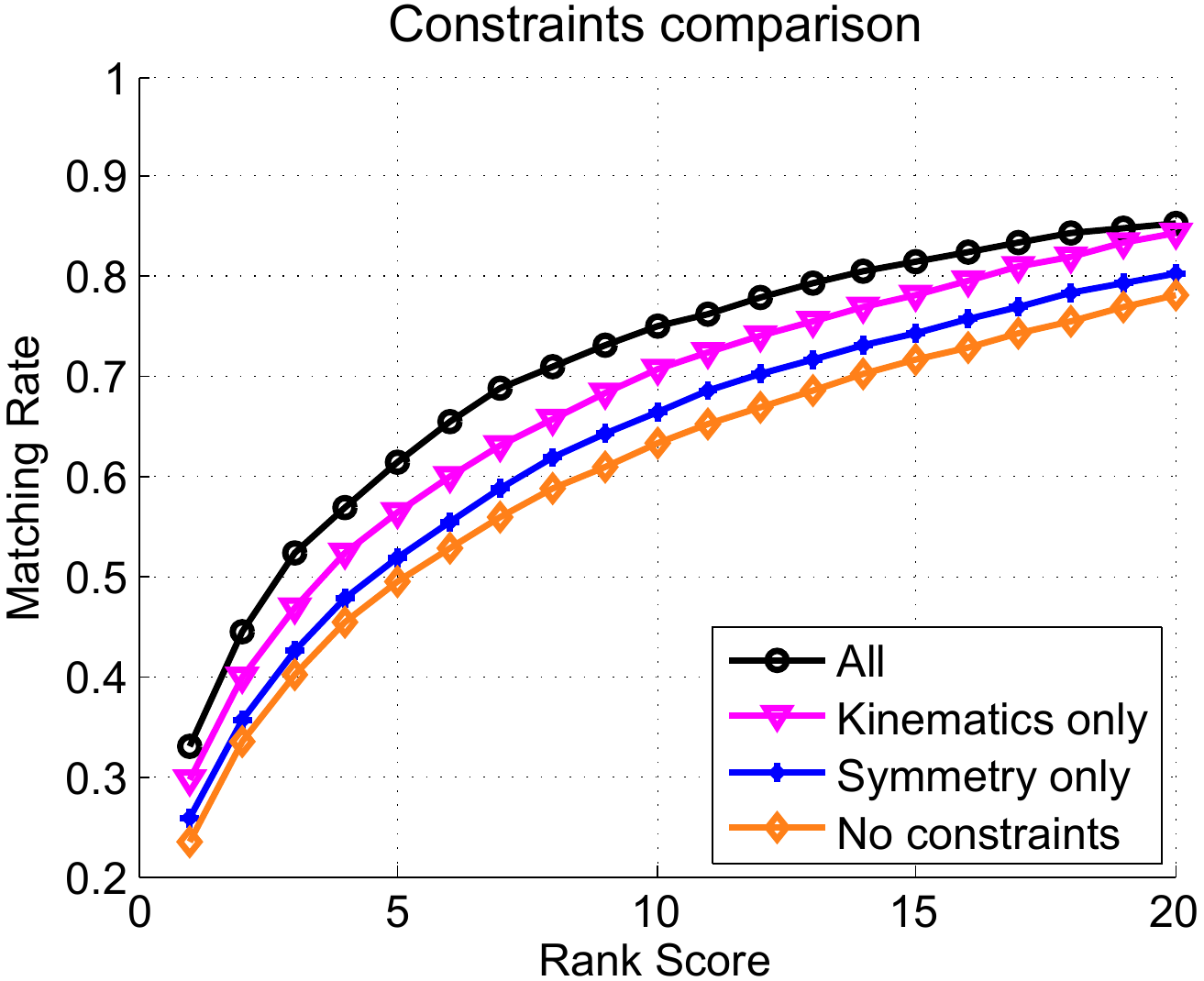}
\end{center}
\vspace{-4mm}
\caption{Empirical studies on different features (left) and constraints (right) used in our approach on CAMPUS-Human dataset.}
\label{fig:termanalysis}
\vspace{-4mm}
\end{figure}

{\em Component Analysis}. We further analyze component benefits of our approach on CAMPUS-Human dataset under the setting: multi-shot $M=3$. Regarding feature effectiveness, we separately evaluate different image features, as shown in Fig.~\ref{fig:termanalysis}(left). It is apparent that the combined feature improves the result. We also demonstrate the effectiveness of the constraints employed, and Fig.~\ref{fig:termanalysis}(right) confirms that both kinematics and symmetry constraints help construct better matching solution.


\vspace{-3mm}


\vspace{-2mm}
\section{Conclusion} \label{sec:Conclusion}
\vspace{-1mm}

This paper studies a novel compositional template for human re-identification, in the form of an expressive multiple-instance-based compositional representation of the query individual. By exploiting reconfigurable compositions and contextual interactions during inference, our method handles well challenges in human re-identification. Moreover, we will explore more robust and flexible part representations and better inter-part relations in future works.



{\small
\bibliographystyle{ieee}
\bibliography{reidbib}
}

\end{document}